\pdfoutput=1
\documentclass[11pt]{article}
\usepackage{emnlp2021}
\usepackage{times}
\usepackage{latexsym}

\usepackage{colortbl}
\usepackage[utf8]{inputenc}
\usepackage[english]{babel}

\usepackage{array, makecell}
\usepackage{amsmath,amsthm}
\usepackage{multirow}
\usepackage{microtype}
\usepackage{booktabs}
\usepackage{subcaption}
\usepackage{enumitem}

\newcommand\blfootnote[1]{%
  \begingroup
  \renewcommand\thefootnote{}\footnote{#1}%
  \addtocounter{footnote}{-1}%
  \endgroup
}
\usepackage{inconsolata}
\usepackage{tikz}

\usepackage[noend]{algorithmic}
\usepackage{algorithm}

\usepackage{microtype}

\title{On the Influence of Masking Policies in Intermediate Pre-training}

\author{Qinyuan Ye$^{1\dagger}$ \quad Belinda Z. Li$^{2\ddagger}$ \quad Sinong Wang$^{3}$ \quad Benjamin Bolte$^{3}$\\\textbf{Hao Ma$^{3}$  \quad Wen-tau Yih$^{3}$ \quad Xiang Ren$^{1}$ \quad Madian Khabsa$^{3}$}\\
$^{1}$University of Southern California \quad $^{2}$MIT CSAIL \quad $^{3}$Facebook AI\\
\texttt{\{qinyuany,xiangren\}@usc.edu} \quad \texttt{bzl@mit.edu}\\
\texttt{\{sinongwang,bbolte,scottyih,haom,mkhabsa\}@facebook.com}
}

\date{}

\begin{document}
\maketitle
\begin{abstract}
Current NLP models are predominantly trained through a two-stage ``pre-train then fine-tune'' pipeline.
Prior work has shown that inserting an \textit{intermediate} pre-training stage, using heuristic masking policies for masked language modeling (MLM), can significantly improve final performance.
However, it is still unclear (1) \textit{in what cases} such intermediate pre-training is helpful, (2) whether hand-crafted heuristic objectives are \textit{optimal} for a given task, and (3) whether a masking policy designed for one task is \textit{generalizable} beyond that task. 
In this paper, we perform a large-scale empirical study to investigate the effect of various masking policies in intermediate pre-training with nine selected tasks across three categories. 
Crucially, we introduce methods to \textit{automate} the discovery of optimal masking policies via direct supervision or meta-learning.
We conclude that the success of intermediate pre-training is dependent on appropriate pre-train corpus, selection of output format (\textit{i.e.}, masked spans or full sentence), and clear understanding of the role that MLM plays for the downstream task.
In addition, we find our learned masking policies outperform the heuristic of masking named entities on TriviaQA, and policies learned from one task can positively transfer to other tasks in certain cases, inviting future research in this direction.
\end{abstract}

\section{Introduction}
\begin{figure}[t]
    \centering
    % \vspace{-0.2cm}
    \includegraphics[width=0.5\textwidth,trim=0 1.3cm 3cm 0,clip]{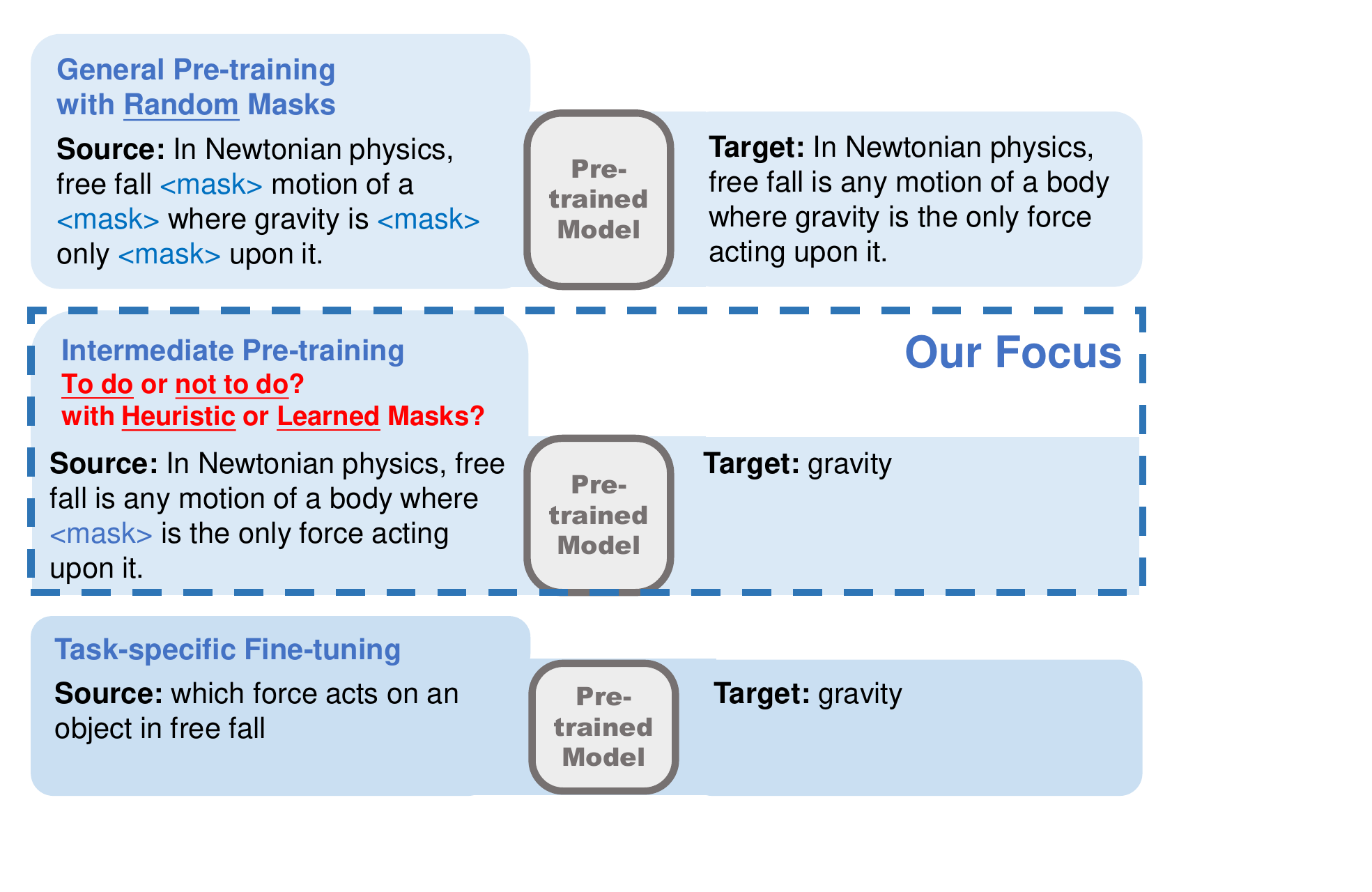}
    % \vspace{-0.6cm}
    \caption{\textbf{Analysis Setup.} We investigate the influence brought by different masking policies during intermediate pre-training, a stage between general pre-training and task-specific fine-tuning. We apply three types of policies (heuristic, supervised, meta-learned) on three categories of tasks (closed-book QA, knowledge-intensive language tasks, multiple-choice QA).}
    \label{fig:intro}
    % \vspace{-0.3cm}
\end{figure}

Large, neural language models (LMs) pre-trained with masked language modeling \cite{devlin-etal-2019-bert,raffel2019exploring} have achieved impressive results over a variety of NLP tasks.\blfootnote{$^\dagger$Work partially done while interning at Facebook AI.}
Studies show that an additional intermediate pre-training stage between general pre-training and task-specific fine-tuning further improves downstream performance (Fig. \ref{fig:intro}).\blfootnote{$^\ddagger$Work partially done while working at Facebook AI.}
For example, intermediate pre-training by masking and recovering named entities or dates, known as salient span masking (SSM, \citealt{guu2020realm}), significantly improves a model's performance of answering factoid questions in a closed-book setting \cite{roberts2020much}.
% For example, salient span masking (SSM), the pre-training objective of recovering masked named entities or dates \cite{guu2020realm}, significantly improves performance of closed-book QA \cite{roberts2020much}, a task that requires a model to answer factoid questions without access to external knowledge. 
% However, it is yet unclear whether these masking heuristics are optimal, or whether they generalize to NLP tasks with different formats and goals.
However, there is a lack of systematic study on how intermediate pre-training works, whether heuristic masking policies like SSM are near-optimal, or whether they generalize to different NLP tasks. Additionally, it is unclear that for tasks other than closed-book QA, whether intermediate pre-training is helpful, or what masking strategy should be adopted. 

In this paper, we offer a large-scale, systematic study on the effects and transferability of masking strategies during intermediate pre-training, while we carefully control all other aspects (\S\ref{sec:analysis-setup}). We first begin our analysis with a focus on three \textbf{heuristic masking policies} (\S\ref{sec:heuristic-policies}). We fine-tune the models resulting from intermediate pre-training on nine selected tasks covering three categories (closed-book QA, knowledge-intensive language tasks, and multi-choice QA). Our results suggest that successful intermediate pre-training is dependent on the selection of appropriate corpus.
Moreover, heuristic-based approaches are effective only when we have a precise understanding of the role masked language modeling (MLM) plays in downstream task. For example, MLM serves as a sort of memorization step \cite{petroni-etal-2019-language}, whereby learning to unmask spans in context is analogous to memorizing facts about the span. In the absence of such understanding, heuristic policies may be sub-optimal.

This motivates us to explore whether automating the discovery of optimal masking policies is possible. We design methods to \textbf{learn a masking policy} with supervised learning (\S\ref{sec:supervised-policy}) or meta-learning (\S\ref{sec:meta-learned-policy}), and compare downstream task performance using the same protocol in our previous analysis. 
Notably, we observe that masking policies learned with supervised learning and meta-learning outperforms the SSM policy for TriviaQA \cite{joshi-etal-2017-triviaqa}, and these policies learned from TriviaQA also help improve performance on Web Questions \cite{berant-etal-2013-semantic}. We also discuss the pros and cons of learned masking policies, such as downstream task learning efficiency, risks of over-fitting and learning instability.

Finally, in hopes to better understand the heuristic and learned masking policies, we provide \textbf{quantitative analysis} on the masks produced by these policies. We visualize the distribution of part-of-speech tags among masked tokens, and their relation to token frequency in the corpus (\S\ref{sec:quant-analysis}). We find that the masking policies learned from TriviaQA tend to mask more proper nouns and tend to mask less frequent words when compared to SSM. 

Overall, our empirical analysis provides useful suggestions for NLP researchers who aim to improve downstream task performance using intermediate pre-training and heuristic masking strategies. In addition, our experiments reveal that infusing task-specific knowledge into LMs with learned masking policies is a promising way to improve downstream task performance, and invite future research in this direction.

\section{Preliminary: Masked Language Modeling}

In this section, we revisit MLM objective with the notation that we will use throughout the paper. MLM is
a predominant pre-training objective for large-scale transformers in NLP
% self-supervised pre-training objective
\cite{devlin-etal-2019-bert}.
MLM and its variants can be characterized with two key components:
a \textbf{masking policy} $g(.;\phi)$, parameterized by $\phi$, which decides the collection of tokens to be masked, and 
a \textbf{language model} $f(.;\theta)$, parameterized by $\theta$.

Formally, given a sequence of tokens $\mathbf{x}=[x_1, x_2, ..., x_m]$, %in the pre-training corpus, 
$g(\mathbf{x};\phi)$ generates a sequence of binary decisions $\mathbf{d}=[d_1, d_2, ..., d_m]$, where $d_i=1$ indicates the token $x_i$ will be masked.
% The tokens $\mathbf{x}$ and the decisions $\mathbf{d}$ will go through a post-processing step, where a source sequence $\mathbf{x}^{(\text{src})}$ and a target sequence $\mathbf{x}^{(\text{tar})}$ are determined.
The source sequence for pre-training, $\mathbf{x}^{(\text{src})}$, is formulated by replacing the selected tokens with a special \texttt{<mask>} token, i.e., $\mathbf{x}^{(\text{src})}=[x^{(\text{src})}_1, x^{(\text{src})}_2, ..., x^{(\text{src})}_m]$, where $x^{(\text{src})}_i=x_i$ if $d_i=0$ and $x^{(\text{src})}_i=\texttt{<mask>}$ if $d_i=1$. We denote this operation as $\mathbf{x}^{(\text{src})} = \mathbf{x} \oplus \mathbf{d}$. The target sequence $\mathbf{x}^{(\text{tar})}$ can be either the full original sequence $\mathbf{x}$ (BART, \citealt{lewis-etal-2020-bart}), or the sequence of masked tokens (T5, \citealt{raffel2019exploring}).
% In salient-span masking (SSM), $g$ is .... In BART pretraining, $g$ is.... \bzl{can we add a sentence about how SSM works (using our notation) here?}

\section{Analysis Setup}
\label{sec:analysis-setup}

% Our goal is to analyze the influence brought by different masking policies during intermediate pre-training. 
% The only variable is the masking policy, while other aspects are controlled. The influence is measured by the performance on downstream tasks. 
In this section we introduce the analysis pipeline (\S\ref{sec:analysis-pipeline}) and downstream datasets we use (\S\ref{sec:downstream-tasks}). 
We defer the details of learned masking policies to \S\ref{sec:compared-policies}.

\subsection{Experiment Procedure} 
\label{sec:analysis-pipeline}
Our goal is to analyze the influence in downstream task performance brought by different masking policies $g(.;\phi)$ during intermediate pre-training. Towards this goal, we ensure that the only variable is the masking policy, while all other aspects are controlled, so that the downstream performance reveal the influence we aim to study.
We first initialize with a BART-base model \cite{lewis-etal-2020-bart};
then for each masking policy, we conduct experiments following a two-stage pipeline: 

\vspace{-0.1cm}
\paragraph{Stage 1. Intermediate Pre-training.} We perform intermediate pre-training with a given masking policy $g(.;\phi)$.
All intermediate pre-training is done with input sequence length of 128, batch size of 2048, learning rate of $0.0001$, up to a total number of $100,000$ updates, using Wikipedia snapshot from December 20, 2018\footnote{The snapshot is available at \url{https://archive.org/download/enwiki-20181220}. Wikipedia is licensed under \href{https://creativecommons.org/licenses/by-sa/3.0/}{CC BY-SA 3.0}.}. 
% The learning rate linearly increase to $0.0001$ in the first 6\% of the updates, and linearly decreases to $0$ in the remaining steps. 
% See Appendix \ref{app:hps-intermediate} for details.

\vspace{-0.1cm}
\paragraph{Stage 2. Task-specific Fine-tuning.} We fine-tune each resulting checkpoint from Stage 1 on related downstream tasks, and evaluate their performance. We follow the same routine of hyperparameter search for each checkpoint. We then run the fine-tuning experiments with the best hyperparameter setting and three different random seeds. See Appendix \ref{app:hps-downstream} for details. 

\subsection{Downstream Tasks and Datasets} 
\label{sec:downstream-tasks}
We focus our study on nine downstream tasks across three categories. We introduce their details and explain the rationale behind our selection in the following.
% \bzl{what if we added a bit of intuition here on what we think the learned masking will do in each case (i.e. in case of cloesd-book qa, it's learning what "knowledge" to pack). idk if it's helpful here or do all analysis post-hoc}

\vspace{-0.1cm}
\paragraph{Closed-book QA.} Closed-book QA is a task that requires a language model to directly answer questions without access to external knowledge \cite{roberts2020much}. This paradigm assumes that the model memorizes large amounts of knowledge from its pre-training data, which gets ``packed" into its parameters, and can subsequently be ``retrieved" to answer questions.
Notably, \citet{roberts2020much} reported 9\%+ improvement in exact match on TriviaQA when intermediate pre-training with salient span masking (\textit{i.e.}, masking and recovering named entities or dates) is performed on a T5-11B model. This observation inspired our work. Our study considers three datasets for closed-book QA: Natural Questions (\textbf{NQ}, \citealt{kwiatkowski-etal-2019-natural}), WebQuestions (\textbf{WQ}) and \mbox{TriviaQA} (\textbf{TQA}).

\vspace{-0.1cm}
\paragraph{Knowledge-Intensive Tasks from KILT.} 
% In order to explore whether salient span masking generalizes to other tasks that requires implicit knowledge stored in the LM, we select three tasks from the KILT benchmark \cite{petroni2020kilt} and conduct experiments in implicit knowledge setting (\textit{i.e.}, without retrieving external knowledge). 
Extending from closed-book QA, we select three tasks from the KILT benchmark \cite{petroni2020kilt} that also aims to test a model's implicit knowledge capacity, while having different task formats and goals.
Aidayago2 (\textbf{AY2}, \citealt{hoffart-etal-2011-robust}) is an entity linking task that requires the model to assign a Wikipedia page to an entity mention in the text. The output is the unique name of the Wikipedia page in text format. 
Zero-shot relation extraction (\textbf{ZSRE}, \citealt{levy-etal-2017-zero}) is a slot filling task that aims to predict the object when given the subject and the relation. The relations in the train/dev/test splits are non-overlapping.
Wizard of Wikipedia (\textbf{WoW}, \citealt{Dinan2019WizardOW}) is a dataset of dialogue histories relevant to knowledge in Wikipedia. The model is required to act like a chatbot and generate the response given previous dialogue history.

% \paragraph{Abstractive Summarization.} Inspired by recent work that pre-trains LMs with gap sentence generation (masking and recovering whole sentences, \citealt{Zhang2019PEGASUSPW}), we explore heuristic and learned policies during intermediate pre-training for abstractive summarization. We use the CNN/Daily Mail (\textbf{CNN/DM}, \citealt{NIPS2015_afdec700}) and Extreme Summarization (\textbf{XSum}, \citealt{narayan-etal-2018-dont}) datasets. For CNN/DM, the summaries are similar to source sequences, while the summaries in XSum are more concise and abstractive.

\vspace{-0.1cm}
\paragraph{Knowledge-Intensive Multiple-choice QA.} 
We select three multiple-choice QA datasets, in which the questions can be answered with commonsense/background knowledge without any context, but the dataset provides additional context paragraphs to explicitly state the background knowledge used. 
% that can be transformed into (question, answer, context) format, where the question can be answered without the context, but the context serves to provide background knowledge. 
We use WIQA \cite{tandon-etal-2019-wiqa} which focuses on procedural text, QuaRTz \cite{tafjord-etal-2019-quartz} which focuses on qualitative relationship, and ROPES \cite{lin-etal-2019-reasoning} which focuses on causes and effects. We reformat these tasks into sequence-to-sequence format, following UnifiedQA \cite{khashabi-etal-2020-unifiedqa}.

% \vspace{0.5cm}
\smallskip
To summarize, all tasks above can be treated as sequence-to-sequence tasks, where each example is a source-target pair $(\mathbf{s}, \mathbf{t})$, accompanied with a context paragraph $\mathbf{c}$ provided by the dataset. Details for dataset splitting are in Appendix~\ref{app:datasets}.
\section{Compared Masking Policies}
\label{sec:compared-policies}
% In this section, we introduce the three categories of masking policies that are compared in this work
We experiment with three categories of masking policies:  %. Namely, we introduce 
heuristic policies, where $g$ is a fixed heuristic function (\S\ref{sec:heuristic-policies}); supervised policies, where $g$ is a model whose weights are learned from \textit{direct supervision} on downstream tasks (\S\ref{sec:supervised-policy}); and meta-learned policies, where $g$ is a model whose weights are learned through \textit{meta-learning} on downstream tasks (\S\ref{sec:meta-learned-policy}).

\vspace{-0.1cm}
\subsection{Heuristic Policy}
\label{sec:heuristic-policies}
\vspace{-0.1cm}
% We take the publicly released BART-base model as the starting point, and include the following heuristic masking policies as baselines:
We experiment with the following three heuristic masking policies:
(1) BART's original denoising objective (\textbf{+Orig});
(2) Masking and recovering 15\% randomly selected tokens (\textbf{+Rand})\footnote{15\% is borrowed from BERT and T5. ``+Orig'' and ``+Rand'' are different in that $\mathbf{x}^{\text{(tar)}}=\mathbf{x}$ for ``+Orig'', while $\mathbf{x}^{\text{(tar)}}$ contains the masked 15\% tokens for ``+Rand''.}; 
(3) Salient span masking, \textit{i.e.}, masking and recovering one named entity \cite{roberts2020much, guu2020realm} (\textbf{+SSM}).

% For summarization tasks, we include two additional heuristics:
% (4) Intermediate pre-training by randomly masking 30\% sentences (+MaskRandomSent);
% (5) Intermediate pre-training by masking the first 30\%\footnote{Borrowed from BART's original masking probability.} sentences (+MaskFirstSent).

\vspace{-0.1cm}
\subsection{Supervised Policy}
\label{sec:supervised-policy}
\vspace{-0.1cm}
When students prepare for closed-book exams, they are likely to review and memorize what they perceive as most important in the text book.  Such perception is learned from their prior experience of taking closed-book exams. Following this intuition, \citet{Ye2020StudyingSL} proposed to learn a masking policy for closed-book QA tasks to help the model focus on \textit{likely answers} during intermediate pre-training. The masking policy is trained with (answer, context) examples, and the policy is an extractive model that extracts the answer span from the context. For example, if the context $\mathbf{x}$ is [Charles, Schulz, was, the, creator, of, Snoopy] and the answer is ``Charles Schulz'', the label for the answer start index will be [1,0,0,0,0,0,0]; for end index it will be [0,1,0,0,0,0,0]. In the following, we briefly recap the method with our notations.

\paragraph{Model.} 
Given context paragraph tokens $\mathbf{x}=[x_1, x_2, ..., x_m]$, we first use an embedding matrix $\mathbf{E}$ to embed each token: $[\mathbf{e}_1, \mathbf{e}_2, ..., \mathbf{e}_n]$. Then, we use a 2-layer bi-directional LSTM model to compute the hidden representation at each position.\footnote{Though the masking policy can theoretically take any form, we opt for a lightweight architecture (2-layer Bi-LSTM) as we need to apply it to millions of pre-training instances.} 
% \vspace{-0.1cm}
\begin{equation}
    [\mathbf{h}_1, \mathbf{h}_2, ..., \mathbf{h}_n] = \textrm{Bi-LSTM}([\mathbf{e}_1, \mathbf{e}_2, ..., \mathbf{e}_n])
\end{equation}

Finally, we use two learned vectors $(\mathbf{w}_{st}, b_{st})$ and $(\mathbf{w}_{ed}, b_{ed})$ to compute the logits for each position being the start or end position of the potential answer/target span. For example, the logit of position $j$ being a start/end position is computed as follows.
% \vspace{-0.1cm}
\begin{equation}
\begin{aligned}
    y_{j, st} = \mathbf{w}_{st} \mathbf{h}_j + b_{st},
    \ \  y_{j, ed} = \mathbf{w}_{ed} \mathbf{h}_j + b_{ed}.
\end{aligned}
\end{equation}

\vspace{-0.3cm}
\paragraph{Policy Inference.} When deploying the policy to intermediate pre-training, we select the potential answer spans by ranking the sum of start and end logits of each potential spans, in accordance to the inference step in machine reading comprehension models. That is, we rank the spans $(i,j)$ according to $y_{i, st}+y_{j, ed}$. We consider two variants when deploying the policy: (a) masking the top 1 span or (b) sampling 1 span from the top 5 spans.

\paragraph{Applicability and Limitation.} Supervised policy is designed for closed-book QA, and one limitation of this method is that the target span $\mathbf{t}$ must appear \textit{as is} in the context paragraph $\mathbf{c}$. Within all other knowledge intensive tasks, only ZSRE satisfies this constraint. To sum up, we apply supervised policy method to TQA, NQ, ZSRE.
% For summarization tasks, we design a workaround by leveraging the extractive summaries of CNN/DM dataset produced by MatchSum \cite{zhong-etal-2020-extractive}. 
% In this case, $\mathbf{c}$ is the original document to be summarized, and $\mathbf{t}$ is one full sentence that appear in the extractive summary. 

\subsection{Meta-learned Policy}
\label{sec:meta-learned-policy}

% \xiang{before going into tech details, say sth about why we want to look at this policy --- any unique characters vs. other policies in comparison?}
Conceptually, what the learned masking policy captures is closely related to the concept of ``learning to learn'' \cite{schmidhuber1987evolutionary, thrun1998learning}. At a high level, the masking policy should provide the model with the desired initialization for the downstream task, such that the model can better learn the downstream task in only a few fine-tuning updates. Therefore, we construct a meta-learning approach, which we describe below. %We formally describe this method below. 
% Additional details can be found in Appendix~\ref{}.%in the following.

\paragraph{Overview.}
We formulate each $(\mathbf{c},\mathbf{s},\mathbf{t})$ example as a small ``task''. For each task, the goal is to improve the performance of generating target sequence $\mathbf{t}$ given input $\mathbf{s}$, immediately after learning from the context $\mathbf{c}$. This is similar to taking quizzes, where a student first learns from a passage $\mathbf{c}$ and then is immediately tested on it by trying to answer $\mathbf{t}$ given $\mathbf{s}$. Studying from $\mathbf{c}$ strategically with an optimal masking policy will result in better performance (\textit{i.e.}, smaller loss in generating $\mathbf{t}$).

\begin{figure}[t]
% \vspace{-0.1cm}
    \centering
    \includegraphics[width=0.5\textwidth,trim=0 1cm 1cm 0.5cm,clip]{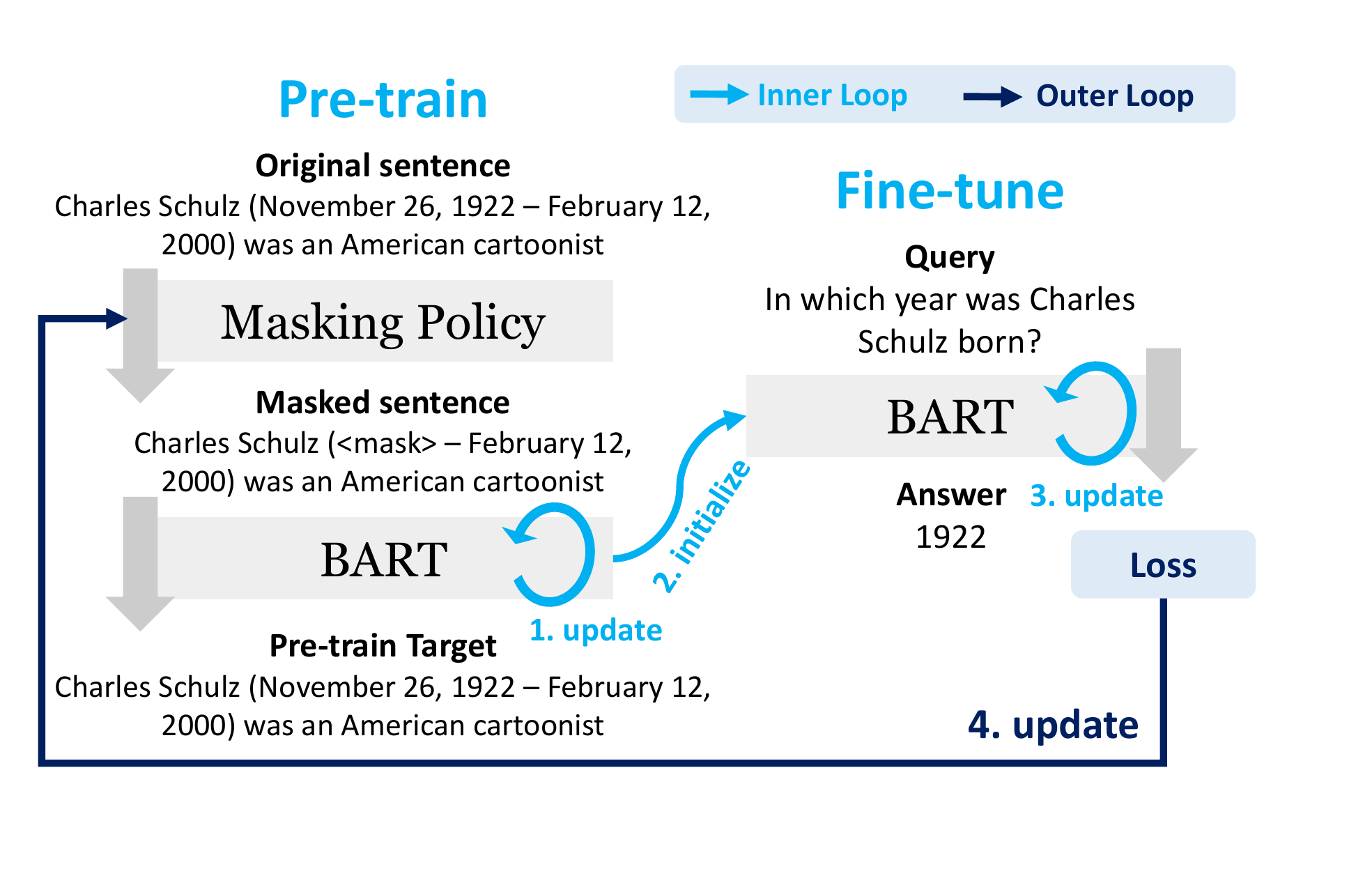}
    % \vspace{-0.7cm}
    \caption{\textbf{Update masking policy by learning from one context-query-answer example.} \textbf{(1) Inner Loop:} (a) A context paragraph $c$ is first masked with current policy $g(.;\phi)$, and the language model is trained to recover masked tokens for one step; (b) the language model is trained on $(q,a)$ pair for one step. \textbf{(2) Outer Loop:} We use the validation loss on the same $(q,a)$ pair to update the masking policy, by directly taking the gradient of loss $L$ w.r.t policy parameters $\phi$.}
    \label{fig:meta-learn}
    % \vspace{-0.3cm}
\end{figure}

Following work in gradient-based meta-learning \cite{pmlr-v70-finn17a, grefenstette2019generalized}, we set up an \textit{inner} and \textit{outer} loop. We briefly sketch the procedure in Fig. \ref{fig:meta-learn}.
In the inner loop, we focus on the current $(\mathbf{c},\mathbf{s},\mathbf{t})$ examples by applying the current masking policy $g(.;\phi)$ and performing pre-train/fine-tune updates to $f(.;\theta)$. In the outer loop, we update the policy $g(.;\phi)$ with the signal at the end of inner loop training. We denote $\phi^{(p)}$ as the masking policy parameters after $p$ outer loop optimization steps, and $\theta^{(p,q)}$ as the LM parameters after $p$ outer loop optimization steps and $q$ inner loop optimization steps.

\paragraph{Inner Loop.} In one inner-loop curriculum, we first take the context as a pre-training sentence, \textit{i.e.}, $\mathbf{x}=\mathbf{c}$, and use the current masking policy $g(.;\phi^{(p)})$ to determine the masks $\mathbf{d}$ and the implied perturbed input $\mathbf{x}^{(\text{src})}$, \textit{i.e.}, $\mathbf{d} = g(\mathbf{x};\phi^{(p)})$, $\mathbf{x}^{(\text{src})} = \mathbf{x} \oplus \mathbf{d}$.
% \begin{align}\label{eq:meta-start}
%     \mathbf{d} = g(\mathbf{x};\phi^{(p)}), \ \ 
%     \mathbf{x}^{(\text{src})} = \mathbf{x} \oplus \mathbf{d}.
% \end{align}

We pre-train $\theta^{(p,0)}$ for one step to recover $\mathbf{x}$ from $\mathbf{x}^{(\text{src})}$:
\begin{equation}\label{eq:meta-start}
    \theta^{(p,1)} = \theta^{(p,0)} - \alpha_0 \nabla_{\theta^{(p,0)}}\mathcal{L}(f(\mathbf{x}^{(\text{src})};\theta^{(p,0)}), \mathbf{x}),
\end{equation}
where $\alpha_0$ is the learning rate, and $\mathcal{L(.,.)}$ is the cross entropy loss for recovering $\mathbf{x}$ using disturbed input $\mathbf{x}^{(\text{src})}$ and model parameters $\theta^{(p,0)}$.

Next, we take $\theta^{(p,1)}$ as initialization and fine-tune it for one step on the downstream objective of predicting $\mathbf{t}$ given $\mathbf{s}$: 
\begin{equation}
    \theta^{(p,2)} = \theta^{(p,1)} - \alpha_1 \nabla_{\theta^{(p,1)}}\mathcal{L}(f(\mathbf{s};\theta^{(p,1)}), \mathbf{t}).
\end{equation}

\paragraph{Outer Loop.} In outer loop, we update the masking policy $g(.;\phi^{(p)})$.
We aim to answer query $\mathbf{s}$ correctly after the inner-loop curriculum. We define the meta-loss $\mathcal{L}'$ as the decrease in losses after the one fine-tuning update, \textit{i.e.},
\begin{equation}\label{eq:meta-end}
    \mathcal{L}' = \mathcal{L}(f(\mathbf{s};\theta^{(p,2)}), \mathbf{t})-\mathcal{L}(f(\mathbf{s};\theta^{(p,1)}), \mathbf{t}).
\end{equation}
$\mathcal{L}'$ characterizes how fast the model has adapted itself to answer ($\mathbf{s}$, $\mathbf{t}$) within one step of optimization.
Since all computations in Eq. (\ref{eq:meta-start}-\ref{eq:meta-end}) are continuous\footnote{We use Gumbel Softmax \cite{Jang2017CategoricalRW} to discretize the output of $g(.;\phi^{(p)})$ and formulate masking decision $\mathbf{d}$, and we use embedding mixture for $\mathbf{x}^{(\text{src})} = \mathbf{x} \oplus \mathbf{d}$}, we optimize $\phi$ by directly taking gradients from $\mathcal{L}'$,
\begin{equation}
    \phi^{(p+1)} = \phi^{(p)} - \alpha_2 \nabla_{\phi^{(p)}}\mathcal{L}'.
    \label{eq:outer_raw}
\end{equation}
% \paragraph{Other Details.} 
% Due to space limit, we leave additional details such as design choices of $g(.;\phi^{(p)})$, controlling masking budget, and post-processing in Appendix \ref{app:meta}.

\paragraph{Controlling Masking Budget.} Higher order optimization is known to be unstable \cite{antoniou2018how}. In early stages of the study, we found the policy to be flipping between masking none or all of the tokens. To stabilize, we add a softened L2 loss to control the portion of mask/not-mask decisions output by $g(.;\phi)$. Denoting $l(\mathbf{x})$ as the input sequence length, $l(\mathbf{d})$ as the number of mask decisions; we define a budget $\gamma$ and a tolerance factor $\epsilon$, and compute the regularization term $\mathcal{L}_{reg}$,
\vspace{-0.15cm}
\begin{equation}
\small
\mathcal{L}_{reg}(\mathbf{x}, \mathbf{d})=\left\{
\begin{aligned}
 & 0, \quad |\gamma l(\mathbf{x}) - l(\mathbf{d})| \leq \epsilon l(\mathbf{x}) \\
%  ((\gamma-\epsilon)l(\mathbf{x}') \leq l(\mathbf{x}'') \leq (\gamma+\epsilon)l(\mathbf{x}')\\
 & (\gamma l(\mathbf{x}) - l(\mathbf{d}))^2, \quad \text{otherwise}
\end{aligned}
\right.
\label{eq:budget}
\end{equation}
% \vspace{-0.15cm}

For example, when $\gamma=15\%$, $\epsilon=5\%$ and the input sequence $\mathbf{x}$ contains 100 tokens, the policy will not be penalized if it's masking $15\pm5$ of all tokens in the sequence. We modify the optimization step in Eq. (\ref{eq:outer_raw}) as follows, where $\beta$ is a co-efficient balancing the regularization intensity.
\begin{equation}
    \phi^{(p+1)} = \phi^{(p)} - \alpha_2 \nabla_{\phi^{(p)}}(\mathcal{L}' + \beta \mathcal{L}_{reg}(\mathbf{x}, \mathbf{d}))\\
\end{equation}

\paragraph{Post-processing.} When we deploy a learned policy to pre-training, we are no longer constrained by differentiability. Based on useful techniques in previous work, we apply post-processing to predicted masking decisions $\mathbf{d}$. (1) Whole-word masking and text infilling \cite{liu2019roberta, lewis-etal-2020-bart}: whenever one subword $x_i$ within a whole word is masked ($d_i=1$), we expand the mask and always mask the whole word. When consecutive tokens are masked, we replace the sequence of \texttt{<mask>} in the input sequence with exactly one \texttt{<mask>} token. (2) Additional budget control: Even with our budget regularization loss (Eq. \ref{eq:budget}), we find some input sequences get too many masks ($>50\%$). This creates extremely challenging pre-train examples that may prevent the model from learning useful information. For these sentences we randomly ``unmask'' tokens to keep the portion of masks below $30\%$.

For space concerns, we leave pseudo-code and other implementation details in Appendix~\ref{app:meta}.

\section{Results and Discussion}

\label{sec:exp-results}
Following our analysis setup (\S\ref{sec:analysis-setup}), we present the results for closed-book QA in Table \ref{tab:cbqa}, knowledge-intensive language tasks (KILT) in Table \ref{tab:kilt} and multiple-choice QA in Table \ref{tab:mcqa}. 
In the following, we aim to understand the influence brought by different masking policies through these results. We also introduce several ad-hoc experiments to verify our hypotheses raised in our analysis.

\subsection{Comparison of Heuristic Policies}
% \smallskip
% \noindent \textbf{Q1. Is intermediate pre-training helpful?} 
\textbf{Continue pre-training with the original objective is helpful in general.} 
Prior work has shown that intermediate pre-training on \textit{encoder} models (\textit{i.e.}, RoBERTa, \citealt{liu2019roberta}) with in-domain corpora helps to improve downstream \textit{classification} tasks performance \cite{gururangan-etal-2020-dont}. 
Our experiments help to examine whether similar conclusion holds for \textit{text-to-text} models and \textit{tasks beyond classification}. 
From our results, we found intermediate pre-training with Wikipedia and BART's original objective (+Orig) improves performance of two closed-book QA tasks (TQA and WQ), one entity linking task (AY2), and two multiple-choice QA tasks (WIQA and QuaRTz); maintains performance on NQ and ZSRE; leads to worse performance on ROPES.
Overall, intermediate pre-training leads to improved performance;
this may be due to the common observation that language models tend to improve with further pre-training even after validation perplexity have plateaued, or that Wikipedia as a general knowledge-intensive corpus, is more closely related to our downstream tasks, compared to the mixture of corpus\footnote{Similar to RoBERTa, BART uses the combination of BookCorpus~\cite{zhu2015aligning}, CC-News~\cite{nagel2016cc}, OpenWebText~\cite{gokaslan2019openwebtext}, and Stories~\cite{trinh2018simple} as pre-training corpus.} used to pre-train BART.

\noindent \textbf{A closer look at ROPES, the exception.} 
We notice that the context paragraphs in ROPES are from science textbooks \textit{and} Wikipedia. We hypothesize that intermediate pre-training on \textit{only} Wikipedia may cause catastrophic forgetting of some scientific knowledge obtained during BART's general pre-training. To verify this, we randomly mask 15\% tokens in ROPES context paragraphs and computed MLM loss. The BART-Base checkpoint achieves 1.97 in NLL Loss, while +Orig achieves 2.02. This supports our hypothesis that intermediate pre-training on a smaller corpus (e.g., Wikipedia) may make the model forget knowledge in general pre-training (e.g., scientific textbooks). We conclude that it is important to pay attention to the corpus from which the downstream dataset is created. 

\begin{table}[t]
\centering
\scalebox{0.7}{
\begin{tabular}{lccc}
\toprule
                        & TQA & WQ & NQ \\
\midrule
BART-Base               & $21.82_{\pm.15}$ & $26.23_{\pm.05}$ & $23.72_{\pm.25}$ \\
\ +Orig                   & $22.91_{\pm.16}$ & \cellcolor{blue!10} $27.17_{\pm.56}$ & \cellcolor{blue!10} $23.85_{\pm.37}$   \\
\ +Rand                   & $22.93_{\pm.14}$ & \cellcolor{blue!10} $27.25_{\pm.68}$ & \cellcolor{blue!10} $24.64_{\pm.44}$   \\
\ +SSM                    & $23.62_{\pm.29}$ & \cellcolor{blue!10} $28.17_{\pm.04}$ & \cellcolor{blue!10} $24.80_{\pm.06}$   \\
\ +Supervised-NQ(Top1)    & $23.48_{\pm.10}$ & \cellcolor{blue!10} $27.43_{\pm.38}$ & \cellcolor{blue!10} $24.58_{\pm.10}$   \\
\ +Supervised-NQ(Top5)    & $23.73_{\pm.21}$ & \cellcolor{blue!10} $28.15_{\pm.05}$ & \cellcolor{blue!30} $24.86_{\pm.28}$   \\
\ +Supervised-TQA(Top1)   & \cellcolor{blue!30}$24.71_{\pm.21}$ & \cellcolor{blue!10} $27.84_{\pm.03}$ & \cellcolor{blue!10} $24.58_{\pm.19}$   \\
\ +Supervised-TQA(Top5)   & $24.43_{\pm.09}$ & \cellcolor{blue!30} $28.35_{\pm.73}$ & \cellcolor{blue!10} $24.66_{\pm.22}$   \\
% \ +Supervsied-All(Top1)   & $24.38_{\pm.14}$ & \cellcolor{blue!10} $28.33_{\pm1.01}$ & \cellcolor{blue!10} $24.55_{\pm.47}$ \\
% \ +Supervsied-All(Top5)   & $24.23_{\pm.25}$ & \cellcolor{blue!10} $28.12_{\pm.36}$ & $24.29_{\pm.17}$   \\
\ +Meta-learned-NQ        & $23.50_{\pm.28}$ & \cellcolor{blue!10} $27.07_{\pm.20}$ & \cellcolor{blue!10} $24.83_{\pm.18}$   \\
\ +Meta-learned-TQA       & $23.88_{\pm.04}$ & \cellcolor{blue!10} $27.49_{\pm.17}$ & \cellcolor{blue!10} $24.85_{\pm.21}$  \\
\midrule
BART-Large & $24.28_{\pm.51}$ & $28.82_{\pm.33}$ & $24.72_{\pm.16}$ \\
\ +Orig                   & $24.34_{\pm.35}$ & $28.28_{\pm.35}$ & \cellcolor{blue!10} $24.91_{\pm.68}$  \\
% \ +Rand                   & - & - & -   \\
\ +SSM                    & $26.29_{\pm.43} $ & \cellcolor{blue!30} $29.79_{\pm.47}$ & \cellcolor{blue!30} $25.34_{\pm.23}$   \\
% \ +Supervised-NQ(Top1)    & - & - & -  \\
\ +Supervised-TQA(Top1)   & \cellcolor{blue!30} $27.18_{\pm.34}$ & \cellcolor{blue!10} $29.71_{\pm.74}$ & $24.28_{\pm.28}$  \\
\bottomrule
\end{tabular}
}
\caption{\textbf{Performance of Closed-book QA Tasks.} We report average and standard deviation of exact match over three runs with different random seeds. {\setlength{\fboxsep}{0pt}\colorbox{blue!30}{Dark blue}} highlights the best performing model. {\setlength{\fboxsep}{0pt}\colorbox{blue!10}{Light blue}} highlights models that are not significantly worse than the best performing model ($p$\textgreater0.1 in paired t-test).}\label{tab:cbqa}
% \vspace{-0.2cm}
\end{table}

\begin{table}[t]
\centering
\scalebox{0.7}{
\begin{tabular}{lccc}
\toprule
                  & AY2 & ZSRE & WoW \\
Metric & EM & EM & F1 \\
\midrule
BART-Base               & $81.07_{\pm.15}$ & $1.89_{\pm.15}$ & \cellcolor{blue!10}$15.14_{\pm.22}$  \\
\ +Orig                   & $81.38_{\pm.06}$ & $1.67_{\pm.15}$ & \cellcolor{blue!10}$15.20_{\pm.13}$  \\
\ +Rand                   & \cellcolor{blue!10}$81.67_{\pm.13}$ & $2.29_{\pm.19}$ & $14.69_{\pm.21}$  \\
\ +SSM                    & \cellcolor{blue!10}$81.74_{\pm.19}$ & \cellcolor{blue!30}$3.52_{\pm.03}$ & $14.68_{\pm.16}$  \\
\ +Supervised-ZSRE(Top1)   & \cellcolor{blue!10}$81.57_{\pm.03}$ & $2.84_{\pm.15}$ & $14.58_{\pm.01}$  \\
\ +Supervised-ZSRE(Top5)   & \cellcolor{blue!30}$81.90_{\pm.22}$ & $2.90_{\pm.03}$ & $14.50_{\pm.38}$   \\
\ +Meta-learned-ZSRE       & $81.31_{\pm.22}$ & $1.99_{\pm.21}$ & $15.07_{\pm.09}$   \\
\ +Meta-learned-WoW        & $80.90_{\pm.23}$ & $1.64_{\pm.05}$ & \cellcolor{blue!30}$15.32_{\pm.05}$   \\
\bottomrule
\end{tabular}
}
\caption{\textbf{Performance of KILT Tasks.}}\label{tab:kilt}
% \vspace{-0.2cm}
\end{table}
% Following \citet{petroni2020kilt}, we evalutate AY2 and ZSRE with Exact Match (EM), and WoW with F1. Coloring scheme is the same as in Table \ref{tab:kilt}.
\begin{table}[t]
\centering
\scalebox{0.7}{
\begin{tabular}{lccc}
\toprule
                  & ROPES & WIQA & QuaRTz \\
\midrule
BART-Base           & $46.60_{\pm0.48}$ & \cellcolor{blue!10}$71.18_{\pm1.12}$ & \cellcolor{blue!10}$62.80_{\pm1.16}$  \\
\ +Orig              & $43.68_{\pm0.67}$ & \cellcolor{blue!30}$73.06_{\pm0.72}$ & $63.35_{\pm0.52}$  \\
\ +Rand                  & $44.59_{\pm1.15}$ & $70.55_{\pm0.42}$ & \cellcolor{blue!10}$63.31_{\pm1.74}$  \\
\ +SSM               & $50.51_{\pm1.15}$ & $69.31_{\pm0.77}$ & \cellcolor{blue!30}$64.41_{\pm1.04}$  \\
\ +Meta-learned-ROPES   & \cellcolor{blue!30}$53.71_{\pm2.33}$ & \cellcolor{blue!10}$73.05_{\pm0.98}$ & \cellcolor{blue!10}$62.93_{\pm1.28}$   \\
\ +Meta-learned-WIQA    & $48.30_{\pm0.69}$ & \cellcolor{blue!10}$72.38_{\pm0.37}$ & \cellcolor{blue!10}$63.14_{\pm1.26}$   \\
\ +Meta-learned-QuaRTz    & $49.01_{\pm1.92}$ & \cellcolor{blue!10}$72.65_{\pm0.53}$ & \cellcolor{blue!10}$63.69_{\pm0.48}$    \\
\bottomrule
\end{tabular}
}
\caption{\textbf{Performance of Multiple-choice QA Tasks.} We report accuracy for each task.}\label{tab:mcqa}
% \vspace{-0.3cm}
\end{table}

\noindent\textbf{Select the heuristic masking policy that resemble the downstream task most.}
So far, we limit the focus to the +Orig objective. Now we further add +Rand and +SSM into the comparison. From the results in Table \ref{tab:cbqa}, we first confirm that salient span masking (SSM) is indeed very beneficial for closed-book QA \cite{roberts2020much}. In addition, SSM helps improve performance for two entity-centric knowledge intensive tasks (AY2 and ZSRE, see Table \ref{tab:kilt}) and two multiple-choice QA tasks (ROPES and QuaRTz, see Table \ref{tab:mcqa}). Note that ROPES focus on causal relationships between entities and QuaRTz focus on qualitative relations (involving numbers); both can be considered entity-centric.
We conclude that using heuristic masking policies that resemble the downstream tasks, or masking information known to be important for the downstream task, tend to improve downstream performance.
When it's difficult to design a heuristic that satisfy these needs, using random masking may be helpful. In this case, we recommend to decide whether to generate full sequence (+Orig) or only masked tokens (+Rand) based on the task output length. If the downstream tasks requires generating long sentences, generating full sequence is more helpful. This is supported by the observation that +Orig is better than +Rand for WoW. On the other hand, if the target sequences in the downstream dataset are shorter, generating masked tokens is more helpful, as shown by experiments on NQ, AY2, ZSRE and ROPES.

% \smallskip
% \noindent \textbf{Q3. Can learned policies be better than heuristics?} 
\subsection{How Do Learned Policies Perform?}
We have introduced two ways to automate the discovery of better masking policies, with supervised learning (\S\ref{sec:supervised-policy}) and meta-learning (\S\ref{sec:meta-learned-policy}). We now extend our analysis to these learned policies.

\noindent \textbf{Successful Cases.} We observe that learned policies are most successful on TriviaQA, with both the supervised policy and the meta-learned policy outperforming SSM. We attribute its success to the following reasons: (1) (context, source, target) examples are abundant, so the masking policy has sufficient supervision. TriviaQA dataset is accompanied with large-scale context paragraphs created with distant supervision, so the scale of $(\mathbf{c}, \mathbf{s}, \mathbf{t})$ examples is larger than other datasets. (2) The heuristic masking policy does not ``perfectly'' resemble the downstream task, and it still has room for improvement. SSM masks one random named entity in the context. However, the answer to trivia questions are not necessarily named entities, and one named entity may be more important than another. 
Therefore the learned policies can better capture the characteristics of TriviaQA than SSM. 
Apart from TriviaQA, meta-learned policies outperforms +Orig on NQ, ZSRE and ROPES, demonstrating the effectiveness of the method. This also opens up a promising direction for downstream tasks whose heuristic masking policy is not intuitive (\textit{e.g.}, dialogue response generation, multiple-choice QA).

\noindent \textbf{Improved learning efficiency.} We additionally consider a low-resource setting for TriviaQA, where we use 0.1\% and 1\% of its training set for fine-tuning. We present the results in Table~\ref{tab:low-resource}. We observe that the supervised policy has better sample efficiency than SSM. We also observe that intermediate pre-training by generating full sequence (a/d) is worse than generating spans (b/c), supporting our previous conclusion that the choice of target sequences should be based on the downstream task output format (span or sentence).

\begin{table}[t]
\centering
\scalebox{0.7}{
\begin{tabular}{lcc}
\toprule
Training Data Used  & 0.1\% & 1\% \\
\midrule
(a) BART-Base               & $3.69\%$ & $5.54\%$\\
(b) +SSM                    & $5.56\%$ & $7.31\%$\\
(c) +Supervised-TQA(Top1)   & $6.49\%$ & $8.40\%$\\
(b) +Meta-learned-TQA       & $4.50\%$ & $6.44\%$\\
\bottomrule
\end{tabular}
}
\caption{\textbf{Performance of TriviaQA in low-resource settings.} Exact match is reported. Supervised policy outperforms other masking policies in low-resource setting, consistent with the full-dataset setting.}\label{tab:low-resource}
% \vspace{-0.2cm}
\end{table}

\noindent \textbf{Overfitting on ZSRE.} ZSRE dataset has a unique setting: it is a slot filling task similar to close-book QA; however it adds additional challenge as the relations in train/dev/test splits are non-overlapping. We hypothesize that this train/test discrepancy leads to unsatisfactory behavior of learned ZSRE policies, and we conduct a set of controlled experiment to validate this hypothesis. Concretely, we use 90\% of its original train set as the new train set, use the 10\% remaining training examples as a ``matched'' dev set, and the original dev set as a ``mismatched'' dev set. In our experiments, SSM achieves $20.02\%_{\pm.16\%}$ EM on match-dev, and $3.21\%_{\pm.15\%}$ on mismatch-dev. Supervised-ZSRE(Top5) achieves $20.37\%_{\pm.04\%}$ on match-dev (outperforms SSM, p$<$0.05), and $2.94\%_{\pm.11\%}$ on mismatch-dev. These experiments show that our supervised policy is learning useful information, but has overfitted to the training data and becomes less robust to distribution shift during inference. 
In comparison, SSM is agnostic to train-test discrepancy and thus achieves the strongest performance.

% Supervised masking policy (\S\ref{sec:supervised-policy}) and meta-learned masking policy (\S\ref{sec:meta-learned-policy}) are our initial attempt towards automating the discovery of optimal masking policy. 

% \smallskip
% \noindent \textbf{Q4. Do masking policies designed for one given task generalize beyond that task?} 
\noindent\textbf{Generalization of learned policies.}
We observe several cases where a policy learned from one dataset positively transfer to another downstream tasks. That includes Supervised-TQA(Top5) bringing improvement to WQ, +Supervised-NQ(Top5) bringing improvement to TQA, and +Supervised-ZSRE(Top5) bringing improvement to AY2, compared to random masking baselines. This is reasonable since all these tasks are entity-centric and are similar in nature. 
For tasks with significantly different formats and goals, \textit{e.g.}, ZSRE and WoW, policies learned on one does not benefit the other. Here we only exhibit the evidence supporting that learned masking policies can positively transfer, and we leave the question of ``when and why does it work'' as future work.

\noindent\textbf{Remarks.} Supervised/meta-learned masking policies are our initial attempt towards the idea of ``learning to mask''. 
While being successful and exhibiting the evidence for positive transfer in certain cases, we recognize the potential risks of over-fitting, or suffering from high instability in meta-learning. 
We hope future work can investigate these issues and design novel methods to learn better masking policies.

\subsection{Quantitative Analysis: What are Masked? How are policies different?}
\label{sec:quant-analysis}
\begin{figure}[t]
    \centering
    % \hspace{-0.7cm}
    \includegraphics[width=0.45\textwidth]{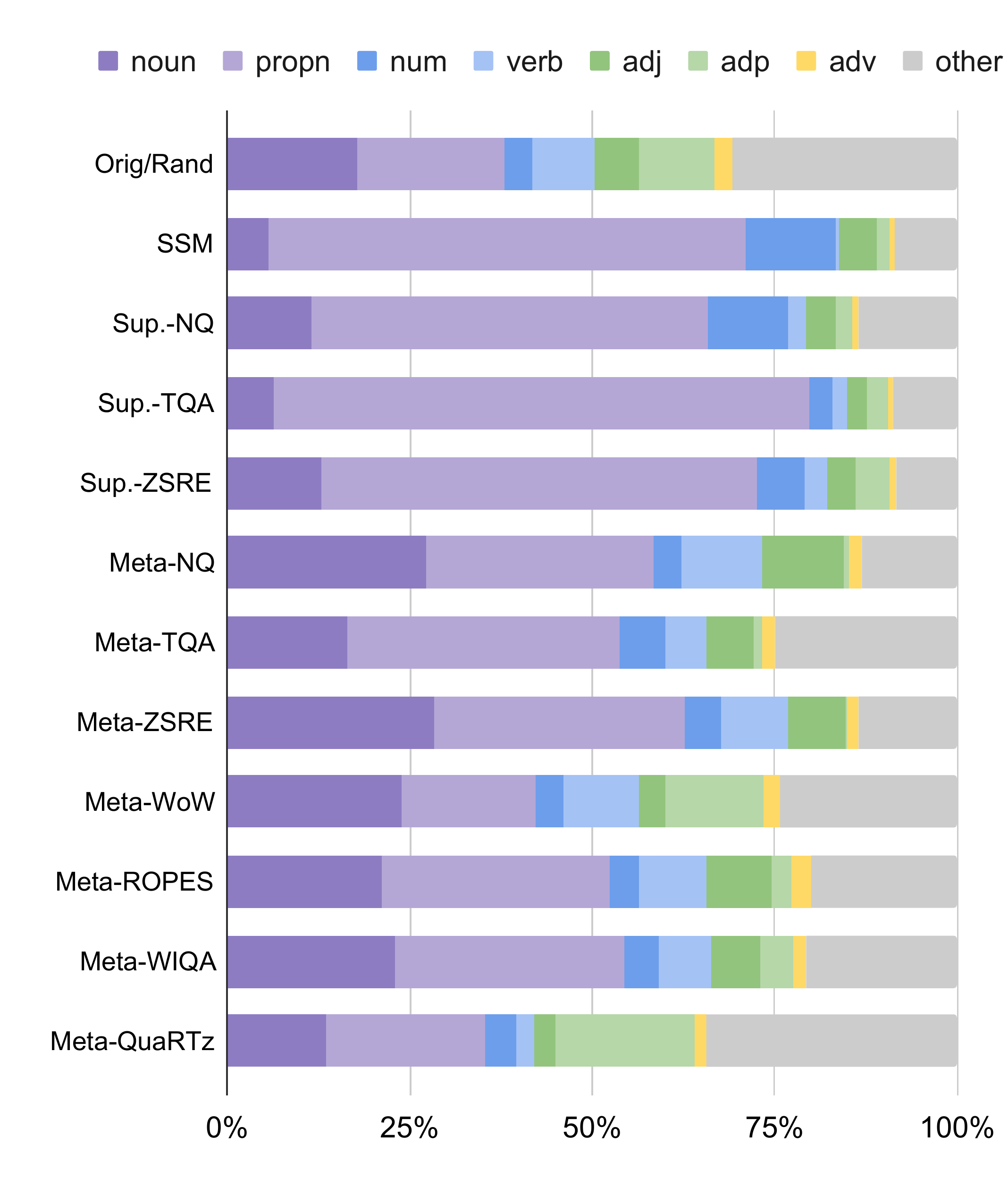}
    \vspace{-0.5cm}
    \caption{Part-of-speech tag distribution for masked produced by different masking policies.}
    \label{fig:pos-tag}
    % \vspace{-0.2cm}
\end{figure}

% \begin{figure}
% \centering
% \begin{subfigure}{.24\textwidth}
%   \centering
%   \includegraphics[width=\textwidth]{figures/pos-tags-1.pdf}
% \end{subfigure}%
% \begin{subfigure}{.24\textwidth}
%   \centering
%   \includegraphics[width=\textwidth]{figures/pos-tags-p2.pdf}
% \end{subfigure}
% \caption{\textbf{Part-of-speech Tags Distribution.}}
% \label{fig:freq}
% \end{figure}
% \textbf{How are these masking policies different from each other?}
In this section we aim to understand how masking policies are different from each other in terms of their masking decisions. We analyze the relation between masking decisions, part-of-speech tags and token frequency. Specifically, we take 1\% of the pre-train corpus and compare the masking decisions made by each policy to facilitate our analysis.

\paragraph{Relation to Part-of-speech Tags.} In Fig. \ref{fig:pos-tag}, we plot the stacked bar chart of part-of-speech tags to visualize their distribution. Each bar represent the portion of masks having the part-of-speech tag, amongst all masks produced by this policy. Notably, most \textit{supervised} policies learns to focus more on proper nouns, and less on common nouns. This is consistent with the goal of the entity-centric downstream tasks. Comparing Supervised-TQA and SSM, Supervised-TQA focuses less on nouns, numbers and adjectives, and it focuses even more on proper nouns. This suggests that Supervised-TQA better characterizes the property of TQA, and thus outperforms SSM by learning to mask task-specific information.
Due to the differences in learning procedures, the \textit{meta-learned} policies has distributions different from supervised policies. Still, meta-learned policies for NQ and TQA masks more proper nouns compared to random masking, similar to their supervised counterparts.
% For QuarRTz, a dataset focusing on qualitative relations, more focus is placed on adpositions (\textit{e.g.}, below, at), and particles (\textit{e.g.}, off, over) compared to other masking policies, which is reasonable
% For WoW, whose heuristic masking policy is less clear, the meta-learned policy tend to mask more nouns and fewer verbs, compared to the random masking baseline.

\begin{figure}
\centering
% \textcolor{blue}{Compiling latex with these two figures will take a lot of time. Putting a placeholder.}
% \vspace{4.5cm}
\begin{subfigure}{.24\textwidth}
  \centering
  \includegraphics[width=\textwidth]{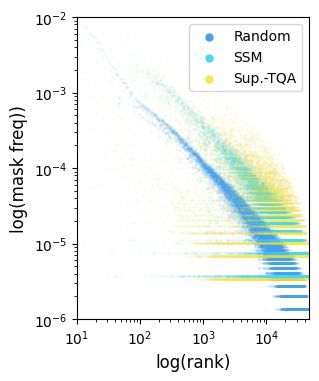}
  \caption{Supervised-TriviaQA}
  \label{fig:sub1}
\end{subfigure}%
\begin{subfigure}{.24\textwidth}
  \centering
  \includegraphics[width=\textwidth]{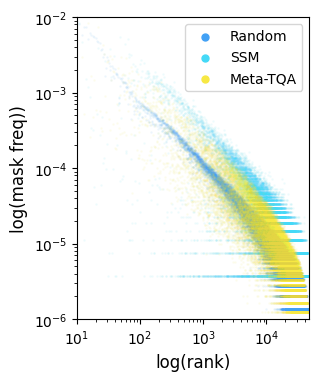}
  \caption{Meta-learned-TriviaQA}
  \label{fig:sub2}
\end{subfigure}
% \vspace{-0.3cm}
\caption{Relation between token frequency rank in the corpus and the token's mask frequency. Best viewed in color.}
\label{fig:freq}
% \vspace{-0.4cm}
\end{figure}

\paragraph{Relation to Token Frequency.} In Fig. \ref{fig:freq}, we plot the relation between mask frequency and token frequency for masking policies learned from TQA, along with random masking and SSM for reference. Mask frequency is computed as the number of occurrences that a token was masked divided by the number of all masked tokens. 
For random masking, the datapoints approximate a Zipfian distribution \cite{zipf1999psycho}, with some noise due to random sampling of words. 
Secondly, for SSM, most datapoints fall on a curve above the random masking line, while a small portion of tokens are less likely to be masked, formulating line segments in the bottom area. These observations indicate that SSM tend to mask less frequent tokens, but its behavior is not fully explained away with token frequency.
The two learned policies, Supervised-TQA (Fig. \ref{fig:freq}(a)) and Meta-TQA (Fig. \ref{fig:freq}(b)) are in general similar to SSM, while the curve for Supervised-TQA is more scattered, indicating a weaker preference for Zipfian behavior.

% % \todo{the frequency figure will slower the latex compilation. deleting for now but will add back.}
% \input{figures/2d_visualization}
% Fig. \ref{fig:2d} visualizes the similarity between different masking policies. We take one minus the average cosine similarities of masking decisions made for pre-train examples as the distance between two policies, and use multi-dimensional scaling algorithm to project to 2D space. We observe that policies learned from TQA and ZSRE are close to SSM in both figures, which is in line with the observation that a majority of the targets in these two datasets are named entities and dates. We also see that policies learned from closed-book QA and knowledge-intensive tasks tend to be close to each other, since both categories examine a model's capacity of implicit knowledge.

% \input{sections/6_experiments_summarization}
% \input{sections/x_related_work}
\section{Related Work}

\paragraph{Implicit Knowledge in Pre-trained Language Models.} 
\citet{petroni-etal-2019-language} discovered that pre-trained language models can implicitly store relational knowledge in their parameters, and such knowledge can be accessed with cloze-style queries. \citet{roberts2020much} introduced the task of closed-book QA, which breaks the convention of retriever-reader strategy for open-domain QA, and requires the model to directly generate answers with its implicit knowledge. Closed-book QA performance is boosted significantly when salient span masking \cite{guu2020realm} is used. \citet{guu2020realm} maintained that SSM helps the model to ``focus on problems that require world knowledge''.

\paragraph{Self-supervised Pre-training.} 
Pre-trained language models has shown its capability on a wide variety of NLP tasks. Current self-supervised objectives are mostly \textit{heuristic}, including masked language modeling \cite{devlin-etal-2019-bert}, span boundary representation learning \cite{joshi-etal-2020-spanbert}, corrupted sentence reconstruction \cite{lewis-etal-2020-bart}, etc. \citet{raffel2019exploring} systematically studied the self-supervised objectives used in previous literature. Related to our goal of exploring pre-training objectives, ELECTRA \cite{clark2020electra} propose a replaced token prediction task which improves pre-training efficiency. \citet{chen2020variance} propose to reduce the variance of gradients in SGD and expedite model pre-training. \citet{Levine2020pmimasking} propose to mask n-grams according to Pointwise Mutual Information (PMI). These works typically consider the efficiency of an objective when pre-training \textit{from scratch} and without preconceived focus on a given problem; while we focus on encoding knowledge or adapting the model during intermediate pre-training with a given task in mind.

\paragraph{Domain/Task-specific Pre-training.}
\citet{gururangan-etal-2020-dont} experiment on four domains (bio-medical, computer science, news, reviews) and eight different datasets, where they discover that pre-training with in-domain corpus leads to better downstream performance.  \citet{kang-etal-2020-neural} propose to learn a mask generator via reinforcement learning. Closely related to us, \citet{gu-etal-2020-train} propose task-guided pre-training by learning to predict importance score for each token in pre-train corpus.
% and then launch task-specific pre-training by masking important tokens.
\citet{vu-etal-2020-exploring, pruksachatkun-etal-2020-intermediate} studies knowledge transfer from intermediate-task fine-tuning, while we focus on a different problem setting of intermediate pre-training with generic corpus (\textit{e.g.}, Wikipedia). We believe both settings have practical utility in real-world applications.

% intermediate-task transfer between NLP tasks. These works performs intermediate  

% \input{sections/7_conclusion}

\section{Conclusion}
In this paper, we study the influence brought by different masking policies used during intermediate pre-training, and offer two methods as our initial attempts towards automating the discovery of optimal masking policy.
From extensive experiments with heuristic and learned masking policies across three categories of tasks, we have identified several successful cases of intermediate pre-training, offered in-depth analysis and insights for the masking policies we used, discussed the risks of learned masking policies, and summarized several suggestions for researchers who wish to adopt intermediate pre-training in their applications.

We also acknowledge that, despite our additional efforts and experiments, several observations still cannot be explained away. We invite future research into this challenging and under-explored problem, to expand on our methods, and to search the space of pre-training objectives beyond masked language modeling. Furthermore, we hope our work encourages researchers to consider the type of downstream applications they wish to deploy their LMs in, before investing resources into large-scale pre-training.

\section*{Acknowledgments} 
We would like to thank anonymous reviewers for their constructive feedback. We would like to thank Sebastian Riedel, Edward Grefenstette, Douwe Kiela, Patrick Lewis, Xian Li, Sewon Min, and USC INK Lab members for the insightful discussions. Qinyuan Ye and Xiang Ren are supported in part by the Office of the Director of National Intelligence (ODNI), Intelligence Advanced Research Projects Activity (IARPA), via Contract No. 2019-19051600007; the DARPA MCS program under Contract No. N660011924033; the Defense Advanced Research Projects Agency with award W911NF-19-20271; NSF IIS 2048211.

\bibliography{anthology,acl2021}
\bibliographystyle{acl_natbib}

\clearpage
\appendix

\section{Additional Training Details}
\subsection{Supervised Policy}
\label{app:supervised}
\paragraph{Training Details.} The embedding matrix $\mathbf{E}$ is initialized with the weights in BART-base model. We optimize cross entropy loss between the logits outputted by the model and the gold annotations. For each source of supervision stated above, we train the policy for 30 epochs with learning rate of 1e-5 and batch size of 512, and select the best checkpoint according to validation loss. 

\subsection{Meta-learned Policy}
\label{app:meta}
{
\let\oldReturn\Return
\renewcommand{\algorithmicrequire}{\textbf{Input:}}
\renewcommand{\algorithmicensure}{\textbf{Output:}}
% \SetAlCapNameFnt{\small}
% \SetAlCapFnt{\small}
\begin{algorithm*}[tb]

\begin{small}
\begin{algorithmic}[1]
\renewcommand{\COMMENT}[2][.5\linewidth]{%
  \leavevmode\hfill\makebox[#1][l]{//~#2}}
\caption{Meta-learning Policy $g(.;\phi)$}\label{algo:overview}
\REQUIRE{Dataset $\mathcal{S}=\{(\mathbf{c}, \mathbf{s}, \mathbf{t})\}$}\;
\ENSURE{Masking Policy $g(.;\phi)$, A Pre-trained Language Model $f(.;\theta)$}\;

% \STATE\textbf{// Stage 1: Meta-learning Policy $g(.;\phi)$}
\FOR{$p = 1 .. T$}
\STATE $\{(\mathbf{c}, \mathbf{s}, \mathbf{t})\} = \text{SampleBatch}(\mathcal{S})$\;
\STATE$\mathbf{x} \mathbf{c}$; $\mathbf{d} = g(\mathbf{x};\phi^{(p)})$; $\mathbf{x}' = \mathbf{x} \oplus \mathbf{d}$ \quad \COMMENT{Apply current masking policy $g(.;\phi)$}\;
\STATE$\theta^{(p,1)} = \theta^{(p,0)} - \alpha_0 \nabla_{\theta^{(p,0)}}\mathcal{L}(f(\mathbf{x}';\theta^{(p,0)}), \mathbf{x})$ \COMMENT{Inner Loop Update 1: Pre-train on $(\mathbf{x}, \mathbf{x}')$}\;
\STATE$\theta^{(p,2)} = \theta^{(p,1)} - \alpha_1 \nabla_{\theta^{(p,1)}}\mathcal{L}(f(\mathbf{s};\theta^{(p,1)}), \mathbf{t})$ \COMMENT{Inner Loop Update 2: Fine-tune on $(\mathbf{s}, \mathbf{t})$}\;
\STATE$\mathcal{L}' = \mathcal{L}(f(\mathbf{s};\theta^{(p,2)}), \mathbf{t})-\mathcal{L}(f(\mathbf{s};\theta^{(p,1)}), \mathbf{t})$ \COMMENT{Compute loss measuring how fast the model adapts to $(\mathbf{s}, \mathbf{t})$}\;
\STATE $\mathcal{L}_{reg} = \delta[|\gamma l(\mathbf{x}) - l(\mathbf{d})| > \epsilon l(\mathbf{x})](\gamma l(\mathbf{x}) - l(\mathbf{d}))^2$ \COMMENT{Compute regularization loss to control masking budget}\;
\STATE $\phi^{(p+1)} = \phi^{(p)} - \alpha_2 \nabla_{\phi^{(p)}}(\mathcal{L}' + \beta \mathcal{L}_{reg}(\mathbf{x}, \mathbf{d}))$ \COMMENT{Outer Loop Update for $\phi$}\;
\STATE $\theta^{(p+1,0)} = \theta^{(p,1)}$ \COMMENT{Maintain pre-train progress at timestamp $p$}\;
\ENDFOR

\end{algorithmic}
\end{small}
\end{algorithm*}
}
\paragraph{Design choices.} We use a 1D convolution layer with two additional linear layers as our policy network $g(.;\phi)$. The linear layers output two logits for each token in input sequence $\mathbf{x}$. The two logits for each tokens go through Gumbel Softmax \cite{Jang2017CategoricalRW} to decide whether it should be masked ($d_i=1$) or not ($d_i=0$). We've also experimented with Bi-LSTM as the encoder, but find meta-learning with LSTMs to be extremely unstable.

\paragraph{Intuitive Example.} The PTLM $f$ is given a piece of context $\mathbf{c}$ ``Charles Schulz (November 26, 1922 – February 12, 2000) was an American cartoonist'' and is expected to take an upcoming ``closed-book exam'' based on this piece of context. In the pre-train step, the current policy $g$ predicts masks (e.g., Charles Schulz (\texttt{<mask>} – February 12, 2000) was an American cartoonist) and take one step of optimization, implicitly encoding this piece of knowledge into its parameters. After this, the PTLM ``transit to closed-book exam mode'' by fine-tuning on $(\mathbf{s}, \mathbf{t})$ for one step. Finally the language model ``takes the closed-book exam'' and the loss for generating $\mathbf{t}$ given $\mathbf{s}$ as input can be interpreted as the supervision for the masking decisions (i.e., whether masking ``November 26, 1922'' is helpful). 

\paragraph{Pseudo-code.} We provide pseudo-code for our method in Algorithm 1.
% \paragraph{Early Stopping Criteria.} There is no direct development set loss we can measure. Moreover, given that gradient-based meta-learning can be unstable, we make early stopping decisions and select checkpoints based on the similarity of its predictions to neighbouring checkpoints. Concretely, if $(c_1, c_2)$ are consecutive checkpoints, we compute their masking decisions on 5 hand-picked sentences, and calculate the $L2$ distance of these predictions (denote as $d(c_1, c_2)$. We calculate this distance for each pair of $(c_i, c_{i+1})$, and select the checkpoint $c_i$ with the lowest $d(c_i,c_{i+1})$. The intuition is that a lower $d(c_i, c_{i+1})$ demonstrate the training is stable and has converged.

\section{Hyperparameters}

% \subsection{Intermediate Pre-training}
% \label{app:hps-intermediate}

% Exceptions are made to pre-training with +MaskFirstSentence, +MaskRandomSentence, +Supervised-CNN/DM, +Meta-learned-CNN/DM, +Meta-learned-XSum. For these intermediate pre-training jobs, we change the input sequence length to 512 to better accommodate summarization tasks.

% \subsection{Downstream Tasks}
For downstream task fine-tuning, we first select the learning rate from $\{$5e-6, 1e-5, 2e-5, 3e-5$\}$ and then fix learning rate to select batch size from $\{$32, 64, 128, 256$\}$. See Table \ref{tab:hps} for more details. 
% For summarization tasks, we first tune the learning rate from $\{$1e-5, 2e-5, 3e-5,5e-5, 1e-4$\}$ and total number of steps in $\{$20000, 30000, 40000, 50000$\}$ with BART-Base. 
% We then continue using the best set of hyperparameters for other checkpoints from intermediate pre-training.

\label{app:hps-downstream}
\begin{table}[h]
\centering
\scalebox{0.7}{
\begin{tabular}{ll}
\toprule
Parameter Name            & Value                          \\
\midrule
Max Epoch                 & 100                            \\
Validation Interval       & 2 or 5                         \\
Warmup Updates            & 500                            \\
Learning Rate             & $\{$5e-6, 1e-5, 2e-5, 3e-5$\}$ \\
Batch Size                & $\{$32, 64, 128, 256$\}$       \\
Label Smoothing           & 0.1                            \\
Dropout                   & 0.1                            \\
Weight Decay              & 0.01                           \\
Clip Norm                 & 0.1                            \\
Generation Beam Size      & 4                              \\
Generation Min/Max Length & 1/20                           \\
Generation Length Penalty & 1.0                           \\
\bottomrule
\end{tabular}
}
\caption{Hyperparameters for Downstream Task Fine-tuning.}\label{tab:hps}
\end{table}

\section{Discussion on NQ} 
\label{app:nq}
In Table \ref{tab:cbqa} we observe that performances on NQ are close for all BART-base models; therefore it is hard to rank all compared methods. We argue that multiple factors leads to this phenomenon, including dataset characteristics and evaluation protocol.
Specifically, NQ may not be an ideal testbed for our study due to three reasons. 

Firstly, intermediate pre-training in general might not be as beneficial for this particular task. For instance,
\citet{roberts2020much} reports only 2\% EM gain on NQ using T5-11B. In our experiments, we use significantly smaller pre-trained models (BART-base/large), so the effect brought by intermediate pre-training will be even smaller. 
In our case we believe the effect is hidden in the variance brought by random seeds. 

Secondly, performance on NQ may not represent the real implicit knowledge capacity of a LM. For reference, we observe a 20\% dev set EM when fine-tuning a randomly initialized BART-base model on NQ. The general pre-training stage brings merely 4-5\% EM improvement, and therefore the improvement brought by intermediate pre-training can be marginal. 

And finally, evaluation based on exact match may substantially underestimate the model capability, as suggested in \cite{roberts2020much}.

\begin{table*}[t]
\centering
\scalebox{0.8}{
\begin{tabular}{p{3.5cm}lccc}
\toprule
Category                                   & Dataset & \#Train & \#Dev & \#Test \\
\midrule
\multirow{3}{*}{\begin{tabular}[c]{@{}l@{}}Closed-book QA\end{tabular}}          & Natural Questions \cite{kwiatkowski-etal-2019-natural}  &  79,168  & 8,757 & 3,610 \\
                                           & WebQuestions \cite{berant-etal-2013-semantic} & 3,417 & 361 & 2,032 \\
                                           & TriviaQA \cite{joshi-etal-2017-triviaqa}      & 78,785 & 8,837 & 11,313 \\
\midrule
\multirow{3}{*}{\begin{tabular}[c]{@{}l@{}}Knowledge-Intensive\\ Tasks (KILT)\end{tabular}} & Aidayago2 \cite{hoffart-etal-2011-robust}       & 18,395 & 4,784 & 4,463  \\
                                           & Zero-shot Relation Extraction \cite{levy-etal-2017-zero} & 147,909 & 3,724 & 4,966 \\
                                           & Wizard of Wikipedia \cite{Dinan2019WizardOW} & 94,577 & 3,058 & 2,944 \\
\midrule
\multirow{3}{*}{\begin{tabular}[c]{@{}l@{}}Knowledge-Intensive \\ Multiple-choice QA\end{tabular}} & ROPES \cite{lin-etal-2019-reasoning} & 10,924 & 844 & 844 \\
                                           & WIQA \cite{tandon-etal-2019-wiqa} & 29,808 & 6,894 & 3,003     \\
                                            & QuaRTz \cite{tafjord-etal-2019-quartz} & 2,696 & 384 & 784    \\
\bottomrule
\end{tabular}
}
\caption{Details of Datasets Used in This Study.}\label{tab:tasks}
\end{table*}

\section{Reproducibility}
\subsection{Dataset Details}
\label{app:datasets}
We obtain closed-book QA datasets from \url{https://github.com/facebookresearch/DPR/blob/master/data/download_data.py}, knowledge-intensive language tasks from \url{https://github.com/facebookresearch/KILT/blob/master/scripts/donwload_all_kilt_data.py}. 
We obtain ROPES, WIQA and QuaRTz from huggingface datasets (\url{https://huggingface.co/datasets}). 
For more details, see Table \ref{tab:tasks}. KILT hosts the test set evaluation on its leaderboard and the test set annotations are not publicly available; therefore we report performance on dev set in Table \ref{tab:kilt}. The test set annotations for ROPES is not publicly available, so we take 50\% of original dev set as the new dev set, and the other 50\% as the new test set.

\subsection{Training Details}
\paragraph{Implementation.} All our experiments are implemented with \texttt{fairseq} \cite{ott-etal-2019-fairseq}. For higher-order optimization in the meta-learning approach optimization, we use \texttt{higher} library \cite{grefenstette2019generalized}. Our code will be released upon acceptance.
\paragraph{Infrastructure and Runtime.} Intermediate pre-training experiments are done with NVIDIA Quadro GP100 or NVIDIA Tesla V100 GPUs, based on availability. For BART-Base, we use 32 GPUs in parallel; For BART-Large, we use 64 GPUs in parallel. Pre-train job takes less than 24 hours for BART-Base models and less than 48 hours for BART-Large models. The checkpoints from intermediate pre-training will be released upon acceptance.
Fine-tuning jobs are all done with one single GPU, with either NVIDIA Quadro GP100, NVIDIA Quadro RTX 8000, NVIDIA Quadro RTX 6000, NVIDIA GeForce RTX 1080 Ti, or NVIDIA GeForce RTX 2080 Ti, based on availability. The list the estimated maximum training time in the following: NQ (4h), WQ (2h), TQA (40h), AY2 (4h), ZSRE (2h), ROPES (1h), WIQA (1h), QuaRTz (1h).
\paragraph{Number of Parameters.} BART-Base model contains 140 million parameters, BART-Large model contains 406 million parameters. Supervised policies contain 43 million parameters (where the word embeddings take 39 millions parameters). Meta-learned policies contain 40 million parameters (where the word embeddings take 39 millions parameters). 

% \section{Full Results of Summarization Tasks}
% \label{app:summ}
% See Table \ref{tab:summarization} for full results (Rouge-1, Rouge-2, Rouge-L) for summarization tasks.

\end{document}